%% file: SPIE2019.tex
\documentclass[]{spie}  %>>> use for US letter paper

\newif\ifnocomment
\nocommentfalse % Show all comments
\ifnocomment
\newcommand{\keymessage}[1]{}
\newcommand{\singchun}[1]{}
\newcommand{\matthias}[1]{}
\else
\newcommand{\keymessage}[1]{\textcolor{blue}{Paragraph Key Message: #1\\}}
\newcommand{\singchun}[1]{\textcolor{green}{SC: #1\\}}
\newcommand{\matthias}[1]{\textcolor{red}{MS: #1\\}}
\fi
\newcommand{\ignore}[1]{}
\usepackage{amsmath,amsfonts,amssymb}
\usepackage{graphicx}
\usepackage{float}
\usepackage[colorlinks=true, allcolors=blue]{hyperref}
\graphicspath{{Fig/}}
\newcommand{\argmin}{\operatornamewithlimits{arg\,min}}
\usepackage{tikz,pgfplots}
\usepackage{tabularx,lipsum,environ}
\usepackage{siunitx}

\newlength{\figureheight}
\newlength{\figurewidth}
\makeatletter
\newcommand{\problemtitle}[1]{\gdef\@problemtitle{#1}}% Store problem title
\newcommand{\probleminput}[1]{\gdef\@probleminput{#1}}% Store problem input
\newcommand{\problemquestion}[1]{\gdef\@problemquestion{#1}}% Store problem
\newcommand{\problemoutput}[1]{\gdef\@problemoutput{#1}}% Store problem oupput
\NewEnviron{problem}{
  \problemtitle{}\probleminput{}\problemquestion{}% Default input is empty
  \BODY% Parse input
  \par\addvspace{.5\baselineskip}
  \noindent
  \begin{tabularx}{\textwidth}{@{\hspace{\parindent}} l X c}
    \multicolumn{2}{@{\hspace{\parindent}}l}{\@problemtitle} \\% Title
    \textbf{Input:} & \@probleminput \\% Input
    \textbf{Problem:} & \@problemquestion \\% Problem
    \textbf{Output:} & \@problemoutput% Output
  \end{tabularx}
  \par\addvspace{.5\baselineskip}
}
\title{Pivot calibration concept for sensor attached mobile c-arms}

\author[a]{*Sing Chun Lee}
\author[b,c]{*Matthias Seibold}
\author[c]{Philipp F\"urnstahl}
\author[d]{Mazda Farshad}
\author[a,b]{Nassir Navab}
\affil[a]{Computer Aided Medical Procedures, Johns Hopkins University, Baltimore, USA}
\affil[b]{Computer Aided Medical Procedures, Technische Universit\"at M\"unchen, Munich, Germany}
\affil[c]{Research in Orthopedic Computer Science, Balgrist University Hospital, University of Z\"urich, Z\"urich, Switzerland}
\affil[d]{Department of Orthopaedics, Balgrist University Hospital, University of Z\"urich, Z\"urich, Switzerland}

\authorinfo{*S.C. Lee and M. Seibold contributed equally to this work and are listed as co-fist authors in alphabetical order.}

\begin{document} 
\maketitle

% Abstract
\input{00-Abstract.tex}

% Introduction
\input{01-Introduction.tex}

% Problem Formulation
\input{02-Formulation.tex}

% Method
\input{03-Method.tex}

% Evaluation
\input{04-Evaluation.tex}

\newpage

% Discussion and conclusion
\input{05-Discussion.tex}

\section{Acknowledgements}
This work is part of the "SURGENT" project under the umbrella of University Medicine Zurich/Hochschulmedizin Zürich.

\newpage

% References
\bibliography{SPIE2019} % bibliography data in report.bib
\bibliographystyle{spiebib} % makes bibtex use spiebib.bst

\end{document}

%% file: 00-Abstract.tex
\begin{abstract}
Medical augmented reality has been actively studied for decades and many methods have been proposed to revolutionize clinical procedures. One example is the camera augmented mobile C-arm (CAMC), which provides a real-time video augmentation onto medical images by rigidly mounting and calibrating a camera to the imaging device. Since then, several CAMC variations have been suggested by calibrating 2D/3D cameras, trackers, and more recently a Microsoft HoloLens to the C-arm.  Different calibration methods have been applied to establish the correspondence between the rigidly attached sensor and the imaging device. A crucial step for these methods is the acquisition of X-Ray images or 3D reconstruction volumes; therefore, requiring the emission of ionizing radiation.
In this work, we analyze the mechanical motion of the device and propose an alternatative method to calibrate sensors to the C-arm without emitting any radiation. Given a sensor is rigidly attached to the device, we introduce an extended pivot calibration concept to compute the fixed translation from the sensor to the C-arm rotation center.  The fixed relationship between the sensor and rotation center can be formulated as a pivot calibration problem with the pivot point moving on a locus.  Our method exploits the rigid C-arm motion describing a Torus surface to solve this calibration problem. We explain the geometry of the C-arm motion and its relation to the attached sensor, propose a calibration algorithm and show its robustness against noise, as well as trajectory and observed pose density by computer simulations.  We discuss this geometric-based formulation and its potential extensions to different C-arm applications. 
\end{abstract}

\keywords{Calibration, Pivot Calibration, Torus, Augmented Reality, Mobile C-arm, CAMC}

%% file: 01-Introduction.tex
\section{INTRODUCTION}
\label{sec:intro}  % \label{} allows reference to this section
% general introduction to the topic of X-ray and CBCT in today orthopedic surgery
Intra-operative X-Ray has become an important tool in orthopedic surgery. Medical experts utilize imaging data acquired by mobile C-arm devices in everyday surgery to position implants, fixate bone fractures, and correct the physiological and mechanical alignment of the skeletal apparatus. Through motorization and calibration, mobile X-Ray systems have been developed to enable intra-operative Cone-Beam Computed Tomography (CBCT) - a reconstruction of 3D volumes from a sequence of X-Ray images along a predefined circular trajectory.%, and hence facilitates intra-operative navigation. %Most C-arm models reconstruct the volume by rotating around a mechanical or virtual axis which defines the reconstruction center of the 3D volume.%SC: This sentence provides extra information but seems not so match with the logic flow.  I think we can elaborate it in the discussion instead

% mention the require of radiation: overlay of X-ray and camera images, or via time consuming calibration procedures
Even though intra-operatively acquired plain radiographs yield great benefits for surgery, the spatial localization and interpretation of 2D X-Ray images is not a trivial task, especially for inexperienced surgeons. For better mental alignment, surgeons often place a radio-opaque object, such as a surgical tool, in the image during its acquisition as an alignment reference. For example, during K-wire placement for a screw insertion in a typical orthopedic surgery, several X-Rays from different perspectives have to be taken to align the K-wire with the desired insertion trajectory. After alignment and insertion, a verification shot is taken to verify the screw position. If the position is not satisfactory, the screw is removed and reinserted which requires the surgeon to perform another mental alignment task. A typical workflow furthermore includes multiple re-positioning of the C-arm machine to restore previous views during surgery. These circumstances lead to high radiation exposure and extended procedure duration. \cite{herscovici2000radiation}%

% introducing navigation system and transit to Medical AR
To leverage the benefits of 3D imaging, CBCT scans for obtaining intra-operative volume data of the patient's anatomy are used in planning and surgical navigation systems. These image guided navigation systems are used to identify anatomical structures intra-operatively and support the surgeon during operation. For conventional navigation systems, a registration step is crucial to align the patient anatomy with the medical data and the surgeon's tools in the tracking coordinate frame. It has been shown that image guided surgery systems have a positive impact on the surgery outcome but can also result in significantly higher operation time. \cite{luz2017imageguided} Augmented Reality (AR) has a great potential to close this gap by providing intuitive medical data visualizations without prolonging the total operation time. For example, the camera augmented mobile C-arm (CAMC) system, which provides intuitive 2D video and X-Ray overlay, has been proposed and showed its promising benefit for surgery. \cite{navab1999merging,navab2010camera,navab2012first} In this paper, we focus on the most crucial step of building CAMC system - the calibration of the attached camera to the imaging device.  In the following section we present an overview of the state-of-the-art calibration methods for sensors and medical imaging systems as applied in the CAMC system. \newline

\subsection{Literature Review}
The idea of the CAMC system is to attach a camera rigidly to the frame of a mobile C-arm and co-calibrate the camera to the X-Ray source to precisely register both image modalities. The original concept proposed by Navab et al. utilized a monocular RGB camera attached to the source of a mobile X-Ray fluoroscopic system. They installed a radio-transparent double mirror construction mounted in the viewing frustum close to the X-Ray source in order to align the optical axes of X-Ray and RGB imaging systems. By modelling the X-Ray source as a pinhole camera and the calibration of both modalities by computing a homography transformation between the two image planes, an overlay of the live camera stream onto the X-Ray image could be accomplished. \cite{navab1999merging} %In addition, to resolve the mechanical instability (vibration of the C-arm during rotation) the calibration is done between the camera to a ``virtual detector plane", and X-Ray is wrapped to the virtual detector plane during use.  %SC: I think ``virtual detector" is solving another problem.  We may just skip this to keep it brief and focus on the calibration part.
As aforementioned, the system was showed to be an intuitive interface for down-the-beam instrument guidance in cadaver studies\cite{navab2010camera} and was used during over 40 orthopedic surgeries in 2012. \cite{navab2012first} Further research on the double-mirror based CAMC system was conducted for projects such as parallax-free image stitching to achieve panoramic X-Ray views. \cite{wang2010parallax} Pauly et al. proposed a learning-based paradigm to identify relevant objects in both X-Ray and optical images to generate a fused image with improved perception of the scene by adjusting the alpha values of the segmented objects in the view. \cite{pauly2014relevance}  % CAMC started the trend of medical AR applications for surgery with a novel design for camera-medical-device calibration.%

The CAMC system was extended by replacing the CCD camera with a depth camera to enable the visualization of X-Ray images projected onto the 3D surface reconstruction of the patient. For calibration, a two-planar calibration phantom was used to align the optical axes of X-Ray and depth camera. Once more, a homography was computed to warp the X-Ray image to the color image plane. \cite{habert2015rgbdx} A similar concept was applied for trackers.  Reaungamornrat et al. presented a system including a tracking camera mounted near the detector of the c-arm. The tracker is calibrated to the flat-panel detector of the mobile x-ray device by acquiring a CBCT scan of a calibration phantom and computing the position of the embedded steel ball bearings in relation to a reference marker. As a result of the calibration, they utilized the hexagonal reference marker to track the gantry during rotation and maintain registration for DRR generation, surgical tracking and video augmentation. \cite{reaungamornrat2011tracking} Albiol et al. calibrated an RGB camera to a conventional diagnostic X-Ray system to identify equivalent points of interest in both modalities using Epipolar geometry. The method enables measurements of real anatomic lengths and angles.  It could serve as inexpensive alternative to CT imaging \cite{albiol2016gcx}.%

In further research projects, the focus shifted from augmenting plain radiographs to create intuitive 3D visualizations using CBCT volume data. A calibration method for the estimation of the transformation between a depth sensor mounted near the detector of the mobile c-arm was proposed by Lee et al. \cite{lee2016calibration} This system enabled a common view of co-registered CBCT volume and real-time point cloud acquired with the depth camera. The calibration method performs the iterative-closest-point (ICP) algorithm to align the volume generated by CBCT with the reconstructed surface of a calibration phantom. An alternative approach utilized a multi-modal chessboard pattern visible in the camera frame and X-Ray image to calibrate the two modalities with a stereo calibration routine. \cite{fotouhi2018augmented} More recently, hand-eye calibration \cite{tsai1989handeye} was applied to calibrate a tracker rigidly mounted on the C-arm gantry for visualizing 3D medical imaging data superimposed on the patient's anatomy \cite{hajek2018closing} and planar X-Ray images in the system's viewing frustum to facilitate the spatial understanding of image acquisition \cite{fotouhi2019iffs}.

% maybe find more external papers

\subsection{Radiation Free Calibration by Analyzing the Geometry}
% Research Gap: tedious calibration procedures (calibration phantoms, stereo calibration, hand-eye,...) --> we use only the geometric properties and the mechanical configuration to calibrate the camera to the ISO center without emitting any radiation!

The above described solutions for calibration use optical axis alignment together with a 2D homography estimation, marker based registration \cite{navab1999merging, habert2015rgbdx}, CBCT acquisition with calibration phantoms\cite{reaungamornrat2011tracking, lee2016calibration}, or hand-eye calibration with the acquisition of up to 160 pose pairs \cite{fotouhi2019iffs}. All presented systems establish correspondences between the two modalities, the attached sensor and the C-arm imaging device by acquiring color and/or depth images and X-Ray or CBCT scans, thus emitting radiation. In the second step, these correspondences are utilized to compute a calibration result, namely a homography or rigid transformation.  

In practice, medical imaging devices are required to be routinely verified for accuracy, and re-calibrated if necessary.  Time-consuming and radiation required calibration methods are not perfectly suitable.  We observed that it is possible to simplify the calibration routine by taking the mechanical configuration of the device into account. In this paper, we propose a radiation free calibration method a trackable sensor to the rotation center of a mobile C-arm device. % The only constraint to this approach is that the C-arm is built with isocenter design and the camera is trackable in world space. 
Our approach exploits the geometrical properties of the mobile C-arm design and the sensor trajectories observed by the tracking system to recover the relation of the sensor's world frame and the C-arm rotation center. %The remaining paper is organized as follows: we describe the mathematical formulation of the problem in Sec.~\ref{sec:problem} and the proposed method in Sec.~\ref{sec:method}. We then present our experiment results in Sec.~\ref{sec:result}, followed by the discussion and conclusion in Sec.~\ref{sec:conclusion}.

%% file: 02-Formulation.tex
\section{PROBLEM FORMULATION}
\label{sec:problem}
The pivot calibration algorithm is applied to recover the translation to the pivot point from poses observed during a spherical motion.  For example, it can be applied for tip position estimation of a tracked tool relative to the reference frame. The pivot point is the center of a sphere and poses are constrained to lie on the sphere surface. This problem can be either solved algebraically using least-square methods or geometrically by sphere fitting.  The well-defined structure of the spherical motion leads to an efficient calibration algorithm. The observation that the typical movement of a mobile C-arm is characterized by a rotation around two different axes leads us to extend the pivot concept to calibrate a sensor to the mobile C-arm rotation center by carefully analyzing the geometry of the C-arm movement and the attached sensor movement.

\subsection{Describing the Mobile C-arm Movement as a Torus}
\begin{figure}[th]
    \centering
    \includegraphics[width=\linewidth]{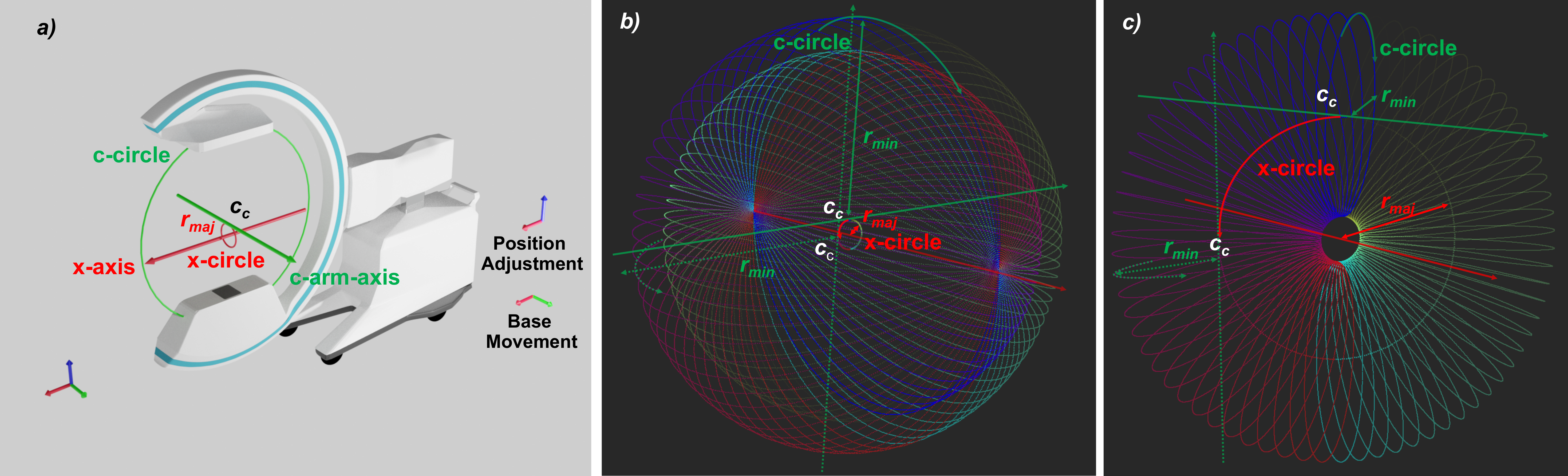}
    \caption{a) illustrates the C-arm movement in terms of its translations and rotations, we denote the movement by rotating around c-arm-axis as c-circle (green), that by x-axis as x-circle (red) and the c-circle center as $c_c$. b) demonstrates the c-circle rotates perpendicularly with $c_c$ on the x-circle.  This forms a locus of a spindle Torus. For comparison, c) shows a standard Torus generated in the same way as in b) with the major radius $r_{maj}$ larger than the minor radius $r_{min}$. When $r_{maj}$ becomes smaller than $r_{min}$, the standard Torus shrinks into a spindle Torus.}
    \label{fig:carm}
\end{figure}

Consider a mobile C-arm in the coordinate system as illustrated in Fig.~\ref{fig:carm}a), its movement has six degrees of freedom, defined by the base movement in the $xy$ plane, position adjustment in the $yz$ plane, and rotations around three axes, namely $x$-axis, $z$-axis and c-arm-axis. If we fix the base movement, the position adjustment, and the $z$-axis rotation, the motion generated by the remaining two, characterized by an offset $r_{maj}$ describes a Torus.

To illustrate this, we denote the rotational motion of the c-arm-axis by ``c-circle" and the one of $x$-axis by ``x-circle", as depicted by the green and red circles/arrows in Fig.~\ref{fig:carm}. The c-circle has the C-arm center $c_{c}$ with radius $r_{min}$ and $c_{c}$ itself rotates around the $x$-axis (red arrows) with radius $r_{maj}$ that forms the x-circle.  We can parameterize the movement of these two circles, $f(\alpha,\beta)$, naturally by $\alpha$ as the amount of rotation in the x-circle and $\beta$ as in the c-circle. $f(\alpha,\beta)$ is then expressed as:
\begin{align}
    \label{eq:torus1}
    f(\alpha, \beta) 
    &= 
    \begin{bmatrix}
    1 & 0 & 0\\
    0 & \cos{\alpha} & -\sin{\alpha} \\
    0 & \sin{\alpha} & \cos{\alpha} \\
    \end{bmatrix}
    \begin{bmatrix}
    \cos{\beta} & 0 & \sin{\beta}\\
    0 & 1 & 0 \\
    -\sin{\beta} & 0 & \cos{\beta}\\
    \end{bmatrix}
    \begin{bmatrix}
    0 \\
    0 \\
    r_{min}
    \end{bmatrix}
    + 
    \begin{bmatrix}
    1 & 0 & 0\\
    0 & \cos{\alpha} & -\sin{\alpha} \\
    0 & \sin{\alpha} & \cos{\alpha} \\
    \end{bmatrix}
    \begin{bmatrix}
    0 \\
    0 \\
    r_{maj}
    \end{bmatrix}
    \\
    \label{eq:torus2}
    &=
    \begin{bmatrix}
    r_{min}\sin{\beta} &
    -(r_{min}\cos{\beta}+r_{maj})\sin{\alpha} &
    (r_{min}\cos{\beta}+r_{maj})\cos{\alpha}
    \end{bmatrix}^T
\end{align}
Eq.~\ref{eq:torus1} describes the movement mathematically by rotating the point $(0,0,r_{min})$ on the c-circle by $\alpha$, then rotating on the x-circle by $\beta$ and translating it by $r_{maj}$ along the rotated axis of the c-circle. After simplifying the equation to Eq.~\ref{eq:torus2}, it in fact is the parametric form of a Torus with a major radius $r_{maj}$ and a minor radius $r_{min}$.  The C-arm movement mentioned above is describing a Torus with the c-circle as its minor circle and the x-circle as its major circle.  Note that $c_{c}$ always lies on the major circle of the Torus - the x-circle.

The surface generated by $f(\alpha,\beta)$ is illustrated in Fig.~\ref{fig:carm}b); however, it may not be obvious to perceive it as a Torus because $r_{maj}$ is considerably smaller than $r_{min}$, which is in fact a special case - namely a ``Spindle Torus".  A standard Torus where $r_{maj}$ is larger than $r_{min}$ is plotted in Fig.~\ref{fig:carm}c) for reference. When $r_{maj}$ becomes smaller than $r_{min}$, the standard Torus shrinks and forms a Spindle Torus like the one in Fig.~\ref{fig:carm}b).  In addition, if $r_{maj}$ is zero, it is degenerated to a sphere with radius $r_{min}$.  In this particular case, $c_{c}$ is stationary.

\subsection{Relations between the Attached Sensor and C-arm Movements}
\label{sec:relationship}
Given a sensor is rigidly attached to the C-arm (i.e. the c-circle), the distance between it and the center $c_{c}$ is constant. If we can pivot around $c_{c}$ directly and collect sensor poses, the offset of the sensor to $c_{c}$ is solvable by standard pivot calibration.  However, we cannot directly pivot the sensor around $c_{c}$, because the movement of the sensor is driven by the C-arm movements and as explained in the previous section, when $r_{maj}$ is non-zero, $c_{c}$ is not stationary but moving on the x-circle.  Our goal is to recover the constant translation from the sensor coordinate system to the rotation center of the imaging system and the torus orientation. Therefore, we need to extend the pivot calibration concept for a moving $c_{c}$ in contrast to a stationary one as in standard pivot calibration. To accomplish this task, we analyzed the geometric relationship between the attached sensor and C-arm movements carefully. Since the distance between the sensor and $c_{c}$ is constant, the offset of the sensor to any given point on the c-circle is also constant.  Take a reference point $(0,0,r_{min})$ on the c-circle and define the offset of the sensor from this reference be $t=(t_x,t_y,t_z)$.  The sensor movement, $g(\alpha,\beta)$, can then be parametrized in the same way as $f(\alpha,\beta)$:
\begin{align}
    \label{eq:etorus1}
    g(\alpha, \beta) 
    &= 
    \begin{bmatrix}
    1 & 0 & 0\\
    0 & \cos{\alpha} & -\sin{\alpha} \\
    0 & \sin{\alpha} & \cos{\alpha} \\
    \end{bmatrix}
    \begin{bmatrix}
    \cos{\beta} & 0 & \sin{\beta}\\
    0 & 1 & 0 \\
    -\sin{\beta} & 0 & \cos{\beta}\\
    \end{bmatrix}
    \begin{bmatrix}
    t_x \\
    t_y \\
    r+t_z
    \end{bmatrix}
    + 
    \begin{bmatrix}
    1 & 0 & 0\\
    0 & \cos{\alpha} & -\sin{\alpha} \\
    0 & \sin{\alpha} & \cos{\alpha} \\
    \end{bmatrix}
    \begin{bmatrix}
    0 \\
    0 \\
    R
    \end{bmatrix}
    \\
    \label{eq:etorus2}
    &=
    \begin{bmatrix}
    (r+t_z)\sin{\beta} + t_x \cos{\beta} \\
    t_y\cos{\alpha} - ((r+t_z)\cos{\beta}-t_x\sin{\beta}+R)\sin{\alpha}\\
    t_y\sin{\alpha} + ((r+t_z)\cos{\beta}-t_x\sin{\beta}+R)\cos{\alpha}
    \end{bmatrix}
\end{align}
Using the fact that the linear combination of sine and cosine functions is the same as a single sine function with a phase shift and scaled amplitude, Eq.~\ref{eq:etorus2} can be re-written as: 
\begin{equation}
    \label{eq:etorus}
    g(\alpha, \beta) = 
    \begin{bmatrix}
    b\sin{(\beta + \gamma)} &
    (b\cos{(\beta+\gamma)}+R)\csc{\delta}\cos{(\alpha + \delta)} &
    (b\cos{(\beta+\gamma)}+R)\csc{\delta}\sin{(\alpha + \delta)}
    \end{bmatrix}^T
\end{equation}
where $b = \sqrt{(r+t_z)^2 + t_x^2}$, $\gamma = \tan^{-1}{\frac{t_x}{r+t_z}}$, and $\delta = \tan^{-1}{\frac{b\cos{\beta+\gamma}+R}{t_y}}$.

\begin{figure}[th]
    \centering
    \includegraphics[width=\linewidth]{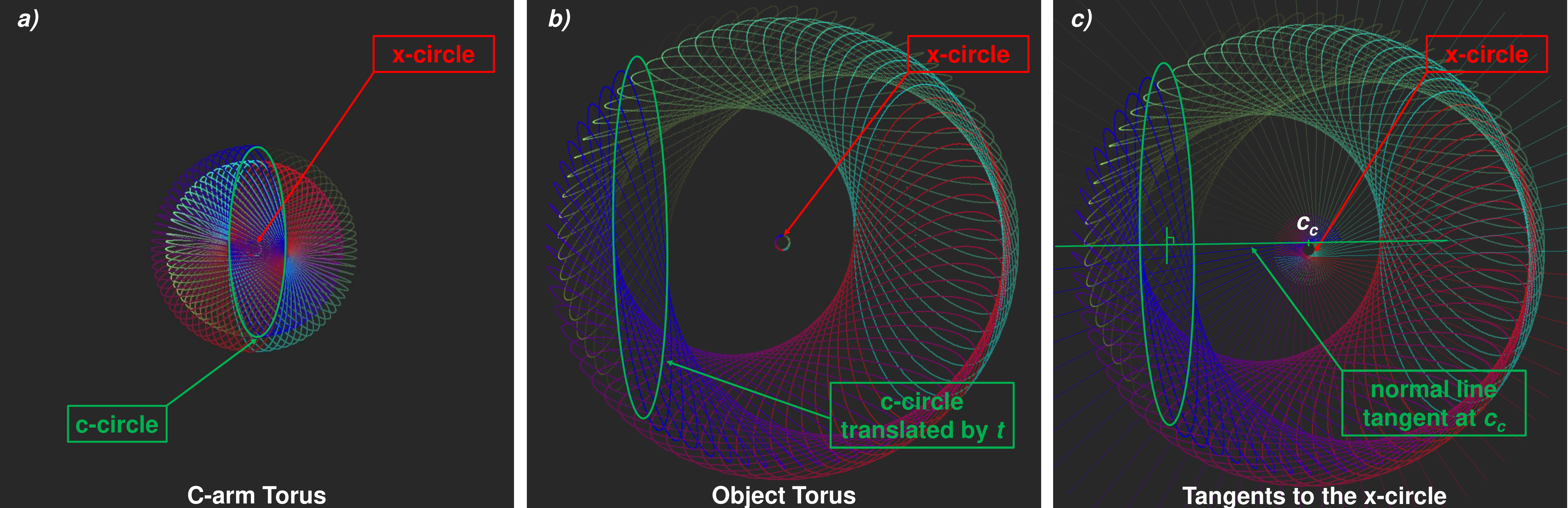}
    \caption{An example of C-arm Torus is shown in a) and the sensor Torus formed due to the offset $t$ is depicted in b).  Since both c-circle and translated c-circle rotate around c-arm-axis, the normal line that passes the center of the translated one tangents the x-circle at $c_{c}$ as illustrated in c).}
    \label{fig:torus}
\end{figure}
% we should make clear that this does not affect the torus shape of the source movement from which we want to extract the major radius ( = tangents of the circular trajectory middle point normals)
Eq.~\ref{eq:etorus} shows that the sensor movement describes a generalized Torus.  An example of the C-arm Torus and the sensor Torus generated due to the positional offset $t$ is visualized in Fig~\ref{fig:torus}a)-b).  Eq.~\ref{eq:etorus} also explains the fact that both Tori share the same center and orientation because their rotation axes are the same.  Furthermore, as depicted in Fig~\ref{fig:torus}c), considering how the sensor Torus is generated, its minor rotation axis coincides with the c-arm-axis.  Therefore, the normal from the sensor Torus' minor center always tangents the x-circle at $c_{c}$. These observations imply that the sensor movement due to the fixed offset $t$ can be seen as \textit{pivoting on the x-circle}.  In a word, we extend the pivot calibration concept in the following way: the sensor (with a fixed offset) is not only pivoting around a stationary point as in the standard pivot calibration, but is pivoting around a point that is moving on a circle.  We call this observed problem the C-arm pivot calibration problem.  % (Note that: the concept can be extended generally to any known locus, which may be useful to restrict the hand-eye calibration problem)

\subsection{C-arm Pivot Calibration}
We define The observed problem as follow:
\begin{problem}
  \problemtitle{\textsc{C-arm Pivot Calibration}}
  \probleminput{Observed poses of the rigidly attached sensor during C-arm movements $\{T_i\}$, where $i$ is the number of observed poses.}
  \problemquestion{Can we recover the offset $t$ of the sensor to the pivot point $c_c$ and the pivot locus (x-circle)?}
  \problemoutput{The locus of $c_{c}$ (the x-circle) and the offset $t$.}
\end{problem}

Since we know the pivot point is moving on a circle, given observed sensor poses $T_i \in SE(3)$, a straightforward way to solve the problem is to solve for a center $c$, a normal $n$, a radius $r_{maj}$, and an offset vector $t$ algebraically by minimizing a 3D circle fitting energy:
\begin{equation}
    \label{eq:pivotenergy}
    \argmin_{c,n,r_{maj},t}\sum_i(||T_i(t)-c|| - r_{maj})^2 + \langle T_i(t)-c,n\rangle^2 \text{ subject to } ||n|| = 1
\end{equation}
where $c$ and $n$ define the pose of the Torus, $r_{maj}$ is the major radius of the Torus and $t$ is offset of the attached sensor to the locus.  Eq.~\ref{eq:pivotenergy} can be solved using non-linear numerical methods such as the Levenberg-Marquardt algorithm. Yet, since pivot calibration is not only solvable in an algebraic way, but also geometrically by fitting a sphere, we believe that C-arm pivot calibration can also be solved in a geometric way.

Based on our analysis of the relation between the C-arm Tours and the sensor Tours in~\ref{sec:relationship}, we can also solve the C-arm pivot calibration problem by fitting the shape using the parametric form in Eq.~\ref{eq:etorus}. However, to apply the calibration result for advanced visualizations of the medical imaging data, we not only need to recover the center but also the orientation of the shape. Therefore, we propose an alternative solution to solve for translation and orientation simultaneously.  We take advantage of the mechanical design of the mobile C-arm, which allows us to observe poses separately for each c-arm-axis and $x$-axis movements and in a defined order (e.g. along the $x-axis$). This enables us to decompose the problem into simpler 2D fitting problems on each observed pose set. Our problem is hence reformulated as follows: 

\begin{problem}
  \problemtitle{\textsc{C-arm Pivot Calibration Reformulated}}
  \probleminput{Observed poses of the rigidly attached sensor during c-arm-axis movements $\{T^c_n\}_m$ and $x$-axis movements $\{T^x_i\}_j$, where $i,n$ and $j,m$ are the number of observed poses and observed sets.}
  \problemquestion{Can we recover the offset $t$ of the sensor to the pivot point $c_c$, the pivot locus (x-circle) and the orientation of the rotation center by fitting shapes in 2D instead of 3D?}
  \problemoutput{The locus of $c_{c}$ (the x-circle), the offset $t$ and the orientation $R$.}
\end{problem}

%\subsection{Relationship to Hand-eye Calibration}

%% file: 03-Method.tex
\section{METHOD}
\label{sec:method}
\begin{figure}[th]
    \centering
    \includegraphics[width=\linewidth]{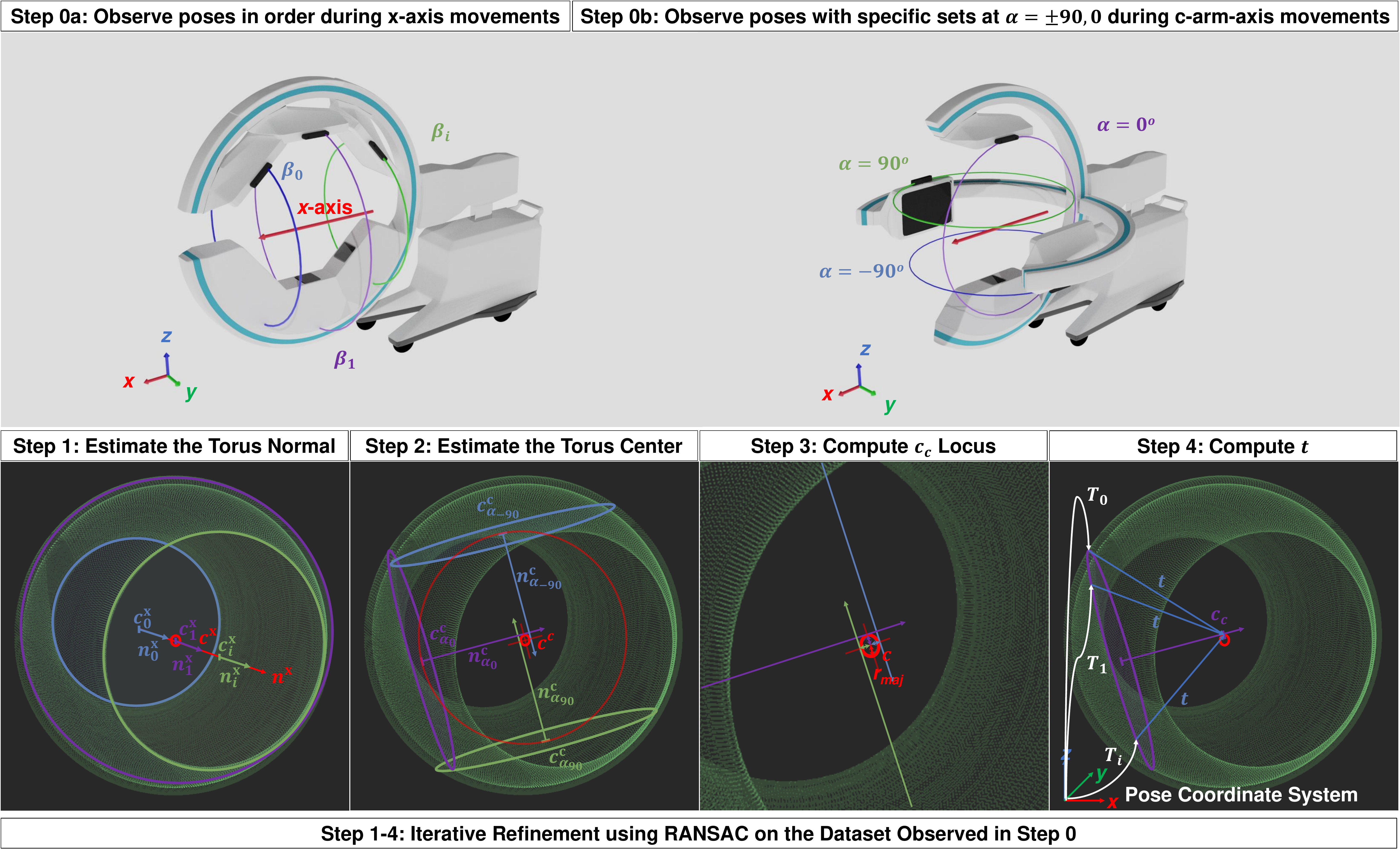}
    \caption{An overview of our method: The input is observed separably at step 0 during the $x$-axis and c-arm-axis movements in a defined order for determining the orientation of the C-arm.  More specifically, this step observes $j$ sets of $x$-axis movements with $i$ poses, $\{T^{x}_i\}_j$, in the order of moving along the positive $x$-axis, and $m$ sets of c-arm-axis movements with $n$ poses, $\{T^{c}_n\}_m$, with specific sets at $\alpha=-90^o,0^o,90^o$.  Step 1 performs circle fittings using $\{T^{x}_i\}_j$ to find the Torus normal $n$ and an center $c^x$, step 2 do the same using $\{T^{c}_n\}_m$ to find another center $c^c$, and the average of $c^x$ and $c^c$ is the estimated Torus center $c$.  The locus of the pivot point $c_c$ is defined by a circle centered at $c$ with the Torus major radius $R$, which is solved by taking the average distance between normal line of the fitted circles in step 2 to $c$.  In step 4, with $\{T^{c}_n\}_m$ at $\alpha = \ang{-90},\ang{0},\ang{0}$, the Torus orientation is deduced and the known $c_c$ at these axes are used together with the corresponding poses to estimate sensor offset $t$.  At last, to increase the robustness to noise, the result is computed iteratively using RANSAC.}
    \label{fig:method}
\end{figure}

Our method decomposes the C-arm pivot calibration problem into multiple 2D geometry problems, which is based on three key observations: 1) the C-arm mechanical design allows us to observe poses separately for the c-arm-axis and $x$-axis movements in a defined order and each movement forms a circle; 2) both the C-arm and sensor Torus share the same center and normal; and 3) the normal line from the minor center of the sensor Torus tangents to x-circle at $c_{c}$. Using these three observations, given input pose sets $\{T^c_i\}_j$ and $\{T^x_i\}_j$ observed during c-arm-axis and $x$-axis movements, we solve the problem in the following steps. These steps are summarized in Fig.~\ref{fig:method}.
\subsection{Estimate the Torus Normal $n$}
By observation 1), $\{T^x_i\}_j$ are poses of $x$-axis movements and each set $j$ observed poses on a circle at a fixed $\beta$.  The center of this circle lies on the $x$-axis and the normal is parallel to it.  During pose observations, we carefully ordered the sets in an ascending order of the $x$-axis so that the order defines the direction of the $x$-axis.  As a result, we can fit a plane on each set to get normal $n^x_j$ (with the same direction as the $x$-axis) and then fit a circle on this plane to get centers $c^x_j$. To increase robustness, we implemented the Random Sample Consensus (RANSAC) algorithm for both plane and circle fittings.  The average of $n^x_j$ is our estimated Torus normal $n$  and the average of $c^x_j$ provides an initial estimate of the center, denoted by $c^x$.

\subsection{Estimate the Torus Center $c$}
Similarly, $\{T^c_i\}_j$ are poses of c-arm-axis movements and each set $j$ contains poses on a circle at a fixed $\alpha$.  As explained earlier, the center and normal of this circle tangents to $c_{c}$ at the x-circle. For each $\{T^c_i\}_j$, we fit a plane and a circle on the plane to get a center $c^c_j$ and a normal $n^c_j$.  The normal direction is deduced by pointing towards $c^x$ found in the previous step. Another circle is fitted to $c^c_j$ with the constraint that the normal is $n$. This circle center $c^c$ is an estimation of the Torus center from $\{T^c_i\}_j$.  By projecting $c^x$ onto this circle gives another estimation of the Torus center mostly from $\{T^x_i\}_j$. We take the average of $c^x$ and $c^c$ as the estimated Torus center $c$.

\subsection{Compute the Pivot Locus of $c_{c}$}
By observation 2), the estimated $c$ is also the center of the C-arm Torus, and by observation 3) we know $n^c_j$ tangents to the x-circle at $c_{c}$. Therefore, the perpendicular distance between $c$ and the line from $c^c_j$ in the direction $n^c_j$ is the radius of the x-circle.  We take their average as the C-arm Torus major radius $r_{maj}$.  The center $c$ and the major radius $r_{maj}$ defines the locus of $c_{c}$.

\subsection{Compute the Local Translation $t$ and the Orientation of the Rotation Center}
\label{method:translation}
To compute $t$, we need to deduce the C-arm Torus orientation. Note that by observation 2) the estimated normal $n$ of the object Torus is the same as the C-arm Torus; therefore, we have estimated the C-arm $x$-axis in the pose observation coordinate system. To compute the remaining two axes, we take advantage of observation 1) to provide as three specific sets that are observed at $\alpha = \ang{-90},\ang{0},\ang{90}$. We project these poses onto the plane with normal $n$ and fit lines to the projected points to recover the two axis directions under the constraint that they are orthogonal. The axis direction is deduced by pointing away from $c$. For these three sets, we also know their $c_{c}$ is on the x-circle intersecting with the computed axis.  We use this information to compute $t$ by averaging: $t = \frac{1}{|\{T_i\}|}\sum_iT_i^{-1}(c_{c})$.

\subsection{Refinement by RANSAC}
At last, because we separate a 3D problem into several 2D sub-problems, the average of the average errors of all sub-problems may not be the same as the average error of the original problem.  To overcome this problem, we apply RANSAC to improve the robustness against noises.

%% file: 04-Evaluation.tex
\section{EVALUATION}
\label{sec:result}
To evaluate the performance of the proposed calibration method, we randomly generated two sets of simulated trajectories and poses with different object translations, torus centers and orientations using Eq.~\ref{eq:etorus2} and averaged our results.  Each data set consists of simulated poses of the c-circle at $\alpha = \ang{-90}, \ang{-80},..., \ang{80}, \ang{90}$ and $\beta = \ang{0},\ang{1},\ang{2},...,\ang{359}$, 
%and of the x-circle at $\alpha = 0^{o},1^{o},2^{o},...,359^{o}$ and $\beta = 0^{o}, 20^{o}, 40^{0}, ..., 360^{o}$, which results in a total number of 13680 simulated poses.  (THIS IS BEFORE DISCARDING THE LAST X-CIRCLE, ALSO WE'RE SIMULATING NOT BETA = 0,20,...,360 BUT 0,20,...,180)
and of the x-circle at $\alpha = \ang{0},\ang{1},\ang{2},...,\ang{359}$ and $\beta = \ang{0}, \ang{20}, ..., \ang{160}$, which results in a total number of 10080 simulated poses.
We conducted three different experiments which showed the proposed method is invariant to the Torus centers and orientations, i.e. independent to the pose observing coordinate system, and examined its performance under noise and insufficient observed data.

\subsection{Performance with Respect to the Torus Pose and Noise Level}
\label{subsec:noise}
In the first experiment, we added white noise to the translation and rotation angles with standard deviations from \SI{0}{\milli\metre} to \SI{5}{\milli\metre} and from \ang{0} to \ang{0.5}, respectively. We estimated the object translation and orientation using our method with 500 RANSAC iterations and \SI{1}{\milli\metre} inlier threshold, and then computed the error between the simulated and estimated translation and orientation. Note that for the orientation error, we compute the angle between rotations by $\theta = \cos^{-1}(\frac{\text{tr}(R)-1}{2})$ where $R = R_{sim}R^{T}_{est}$, $R_{sim}$ and $R_{est}$ are the simulated and estimated orientations.

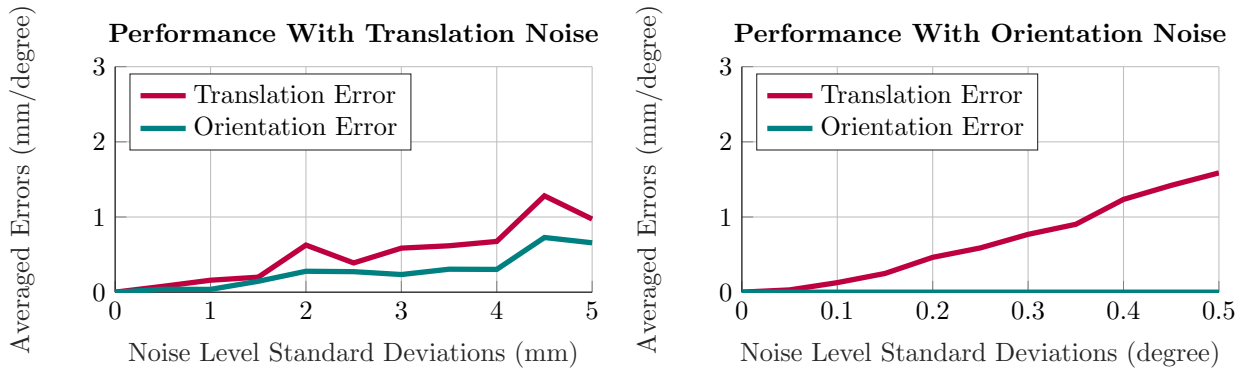
\begin{figure}
    \centering
    \setlength\figurewidth{0.9\textwidth}
    \setlength\figureheight{3 cm}
    \input{Tex/experiment1.tex}
    \label{fig:resultnoise}
    \caption{Performance of the calibration method with increasing translation (left) and orientation noise (right)}
\end{figure}

% As the noise increases, we used the error function, $\text{erf}(\frac{x}{\sqrt{2}})$, where $x$ is the number of standard deviation, to approximate the number of inliers required in the dataset that is within 1 mm inlier threshold.  For the translation noise experiment, we set the number of inliers required to 90\% of the error function value.  
Figure \ref{fig:resultnoise} shows that no additional noise results in an average error of zero which shows that our method is invariant to the pose observation coordinate system. When adding Gaussian noise with a standard deviation of \SI{5}{\milli\metre}, only 20\% of the data lies within \SI{1}{\milli\metre} error. In this case, the method still achieves a comparably small translation error of about \SI{1}{\milli\metre} and orientation error of \ang{0.8}.  
% For the rotation noise experiment, because the Torus shape is randomly generated, there is no direct estimation on how the added rotation noise affects how many inliers are within 1 mm threshold.  In general, if the magnitude of the translation is large, the rotation noise has a bigger effect on the inliers.  To that, we customized the number of inliers required for each noise level.  In practice, this is also a general problem of using RANSAC that a good estimation of this number is required. 
The estimated orientation is independent from the rotation noise because our method estimates the orientation by fitting lines on the estimated plane from pose positions. However, the translation error is affected by orientation noise because the position of the estimated rotation center is computed in respect to the observed poses (see \ref{method:translation}). The method maintains average of under \SI{2}{\milli\metre} error for rotational noise with \ang{0.5} standard deviation.  The RANSAC fitting errors and the number of detected inliers within \SI{1}{\milli\metre} threshold are shown in Tab.~\ref{tab:resultnoise}.  
% Since the rotation noise has bigger effect on the fitting results, in practice, we recommend to capture more poses when it is known that the observed poses contain bigger rotational errors.

\begin{table}[h]
    \centering
    \begin{tabular}{|c||c|c|c|c|c|c|}
        \hline
        \textbf{Translation Noise Standard Deviation (mm)} & \textbf{0} & \textbf{1} & \textbf{2} & \textbf{3} & \textbf{4} & \textbf{5}\\\hline
        Number of Inliers Required & 8064 & 3088 & 1592 & 1067 & 802 & 642\\\hline
        Number of Inliers Detected & 10080 & 3829 & 1953 & 1329 & 1008 & 755\\\hline
        Average Error of Inliners (mm) & 0.0000 & 0.2436 & 0.2465 & 0.2487 & 0.2514 & 0.2549\\\hline
        Average Error of All (mm) & 0.0000 & 0.8008 & 1.6007 & 2.4049 & 3.1928 & 4.2514\\\hline\hline
        \textbf{Rotation Noise Standard Deviation (degree)} & \textbf{0} & \textbf{0.1} & \textbf{0.2} & \textbf{0.3} & \textbf{0.4} & \textbf{0.5}\\\hline
        Number of Inliers Required & 5040 & 1930 & 995 & 667 & 502 & 402\\\hline
        Number of Inliers Detected & 10080 & 2623 & 1333 & 908 & 700 & 561 \\\hline
        Average Error of Inliners (mm) & 0.0000 & 0.2469 & 0.2465 & 0.2459 & 0.2502 & 0.2456\\\hline
        Average Error of All (mm) & 0.0000 & 1.3095 & 2.6175 & 3.9236 & 5.2280 & 6.5317\\\hline
    \end{tabular}
    \caption{This table shows some RANSAC parameters and the fitting errors.}
    \label{tab:resultnoise}
\end{table}

\subsection{Performance with Respect to Number of Observed Trajectories}
\label{subsec:trajectories}
In this experiment, we examined the effect of the number of observed individual trajectories on the performance of our calibration method.  % We generated simulated pose trajectories with $\alpha = -90^{o},-90^{o}+s_{\alpha},-90^{o}+2s_{\alpha},...,90^{o}$ and $\beta = 0^{o},1^{o},2^{o},...,359^{o}$, and of the x-circle at $\alpha = 0^{o},1^{o},2^{o},...,359^{o}$ and $\beta = 0^{o}, s_{\beta},2s_{\beta},...,360^{o}$, where $s_{\alpha}$ and $s_{\beta}$ are step sizes.
For detailed analysis we varied the number of simulated pose trajectories for both c-circles and x-circles while changing the number of one and keeping the other constant to experiment \ref{subsec:noise}.. Note that the minimal number of observed sets are the specific c-circle trajectories at $\alpha = \ang{-90}, \ang{0}, \ang{90}$ and one x-circle trajectory for determining the orientation of the estimated rotation center.  % We reused the simulated dataset and kept decreasing the number of sampling sets of c-arm-axis movements and x-axis movements.  We also decreased the number of sampling in each set and compared their results at noise level with standard deviation 0, 1 for translation and 0, 0.1 for orientation.
For comparability, we fixed the translation noise to \SI{1}{\milli\metre} and the orientation noise to \ang{0.1} and used the same parameters for RANSAC as in the first experiment.

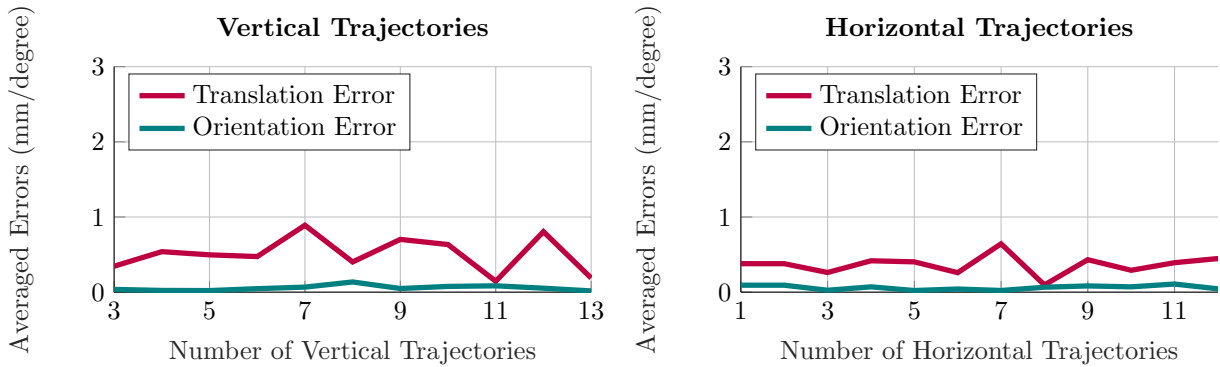
\begin{figure}
    \centering
    \setlength\figurewidth{0.9\textwidth}
    \setlength\figureheight{3 cm}
    \input{Tex/experiment2.tex}
    \caption{Performance of the calibration method with increasing number of observed vertical trajectories (c-circles, left) and horizontal trajectories (x-circles, right).}
    \label{fig:resulttrajectories}
\end{figure}

Both plots in figure \ref{fig:resulttrajectories} show that the performance of the calibration method is not influenced by the number of observed vertical and horizontal trajectories as both rotation and orientation errors do not benefit from increasing the number of trajectories. Therefore, the method is able to give correct results with the minimal observable number of three c-circle and one x-circle trajectories.  

\subsection{Performance with Respect to Changing Density of Poses}

\begin{figure}[H]
    \centering
    \setlength\figurewidth{0.45\textwidth}
    \setlength\figureheight{3 cm}
    \input{Tex/experiment3.tex}
    \caption{Performance of the calibration method with changing density of observed poses.}
    \label{fig:resultdensity}
\end{figure}
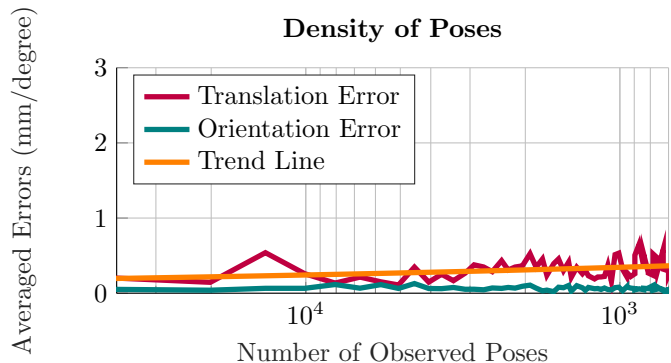

In the last step, we varied the density of observed poses while maintaining the number of observed trajectories, translation and rotation noise, as well as RANSAC parameters equal to experiment \ref{subsec:noise}. The result, illustrated in figure \ref{fig:resultdensity}, shows that decreasing the pose density does not decrease the performance of our algorithm. Furthermore, we investigated the performance with increasing density of poses. It can be observed that the averaged error stays within the bounds of \SI{0}{\milli\metre} to \SI{1}{\milli\metre} and \ang{0} to \ang{1} for translation and orientation, respectively. However, the trend line indicates a slight increase in averaged errors with decreasing pose density. This is due to the fact that by reducing the number of observed data points for c- and x-circles least square fits are becoming more imprecise. Note that the minimal number of points for fitting a plane or a circle is three. %However, in a real world calibration scenario we will use a tracking system to observe the poses of the attached tracker which will result in a sufficient number of observed poses for to robustly estimate the circle and plane geometries. 

%% file: Tex/experiment1.tex
\begin{tikzpicture}

\begin{axis}[%
width=0.411\figurewidth,
height=\figureheight,
at={(0\figurewidth,0\figureheight)},
scale only axis,
xmin=0,
xmax=5,
xlabel style={font=\color{white!15!black}},
xlabel={Noise Level Standard Deviations (mm)},
ymin=0,
ymax=3,
ylabel style={font=\color{white!15!black}},
ylabel={Averaged Errors (mm/degree)},
axis background/.style={fill=white},
title style={font=\bfseries},
title={Performance With Translation Noise},
axis x line*=bottom,
axis y line*=left,
xmajorgrids,
ymajorgrids,
legend style={at={(0.03,0.97)}, anchor=north west, legend cell align=left, align=left, draw=white!15!black}
]
\addplot [color=purple, line width=2.0pt]
  table[row sep=crcr]{%
0	2.31814567541733e-13\\
0.5	0.0772703263529442\\
1	0.15884466994439\\
1.5	0.19926588313039\\
2	0.626590010448915\\
2.5	0.388792149890074\\
3	0.585687320025272\\
3.5	0.617198362181782\\
4	0.674958005133599\\
4.5	1.28182061948665\\
5	0.97232534898042\\
};
\addlegendentry{Translation Error}

\addplot [color=teal, line width=2.0pt]
  table[row sep=crcr]{%
0	0\\
0.5	0.0378632470132523\\
1	0.0371691663816556\\
1.5	0.144778856391094\\
2	0.278083243137815\\
2.5	0.272805088790031\\
3	0.234426747080382\\
3.5	0.305444690526368\\
4	0.302669787110294\\
4.5	0.727293749256744\\
5	0.65685295966621\\
};
\addlegendentry{Orientation Error}

\end{axis}

\begin{axis}[%
width=0.411\figurewidth,
height=\figureheight,
at={(0.54\figurewidth,0\figureheight)},
scale only axis,
xmin=0,
xmax=0.5,
xlabel style={font=\color{white!15!black}},
xlabel={Noise Level Standard Deviations (degree)},
ymin=0,
ymax=3,
ylabel style={font=\color{white!15!black}},
ylabel={Averaged Errors (mm/degree)},
axis background/.style={fill=white},
title style={font=\bfseries},
title={Performance With Orientation Noise},
axis x line*=bottom,
axis y line*=left,
xmajorgrids,
ymajorgrids,
legend style={at={(0.03,0.97)}, anchor=north west, legend cell align=left, align=left, draw=white!15!black}
]
\addplot [color=purple, line width=2.0pt]
  table[row sep=crcr]{%
0	2.32258656751583e-13\\
0.05	0.0300518763126014\\
0.1	0.126728599472506\\
0.15	0.248900679039121\\
0.2	0.463294898414266\\
0.25	0.588788702645588\\
0.3	0.769159184425494\\
0.35	0.902837497119033\\
0.4	1.23412750914059\\
0.45	1.42076218630519\\
0.5	1.58691476552026\\
};
\addlegendentry{Translation Error}

\addplot [color=teal, line width=2.0pt]
  table[row sep=crcr]{%
0	0\\
0.5	0\\
};
\addlegendentry{Orientation Error}

\end{axis}

%\begin{axis}[%
%width=1.227\figurewidth,
%height=1.227\figureheight,
%at={(-0.16\figurewidth,-0.135\figureheight)},
%scale only axis,
%xmin=0,
%xmax=1,
%ymin=0,
%ymax=1,
%axis line style={draw=none},
%ticks=none,
%axis x line*=bottom,
%axis y line*=left,
%legend style={legend cell align=left, align=left, draw=white!15!black}
%]
%\end{axis}
\end{tikzpicture}%

%% file: Tex/experiment2.tex
\begin{tikzpicture}

\begin{axis}[%
width=0.411\figurewidth,
height=\figureheight,
at={(0\figurewidth,0\figureheight)},
scale only axis,
xmin=3,
xmax=13,
xtick={1,3,5,7,9,11,13},
xlabel style={font=\color{white!15!black}},
xlabel={Number of Vertical Trajectories},
ymin=0,
ymax=3,
ylabel style={font=\color{white!15!black}},
ylabel={Averaged Errors (mm/degree)},
axis background/.style={fill=white},
title style={font=\bfseries},
title={Vertical Trajectories},
axis x line*=bottom,
axis y line*=left,
xmajorgrids,
ymajorgrids,
legend style={at={(0.03,0.97)}, anchor=north west, legend cell align=left, align=left, draw=white!15!black}
]
\addplot [color=purple, line width=2.0pt]
  table[row sep=crcr]{%
3	0.345277713973339\\
4	0.539039622774572\\
5	0.496524955345755\\
6	0.473647627059508\\
7	0.89022927786006\\
8	0.401890834433249\\
9	0.702188278646993\\
10	0.633224418605462\\
11	0.145292527612691\\
12	0.805030027153277\\
13	0.188029207850324\\
};
\addlegendentry{Translation Error}

\addplot [color=teal, line width=2.0pt]
  table[row sep=crcr]{%
3	0.0367689444571102\\
4	0.0229349318159322\\
5	0.0210560812445202\\
6	0.0463518609365838\\
7	0.0672319641193866\\
8	0.135584600618591\\
9	0.0484450387038891\\
10	0.0764657629956975\\
11	0.0844458756377424\\
12	0.0526718807789361\\
13	0.016551660232162\\
};
\addlegendentry{Orientation Error}

\end{axis}

\begin{axis}[%
width=0.411\figurewidth,
height=\figureheight,
at={(0.54\figurewidth,0\figureheight)},
scale only axis,
xmin=1,
xmax=12,
xtick={1,3,5,7,9,11},
xlabel style={font=\color{white!15!black}},
xlabel={Number of Horizontal Trajectories},
ymin=0,
ymax=3,
ylabel style={font=\color{white!15!black}},
ylabel={Averaged Errors (mm/degree)},
axis background/.style={fill=white},
title style={font=\bfseries},
title={Horizontal Trajectories},
axis x line*=bottom,
axis y line*=left,
xmajorgrids,
ymajorgrids,
legend style={at={(0.03,0.97)}, anchor=north west, legend cell align=left, align=left, draw=white!15!black}
]
\addplot [color=purple, line width=2.0pt]
  table[row sep=crcr]{%
1	0.378898494219428\\
2	0.378898494219428\\
3	0.261464465217948\\
4	0.417419061025443\\
5	0.403300420343005\\
6	0.259127138933652\\
7	0.643007018528992\\
8	0.0959392106141941\\
9	0.431805965027674\\
10	0.291022390765511\\
11	0.392039955747427\\
12	0.447203659672368\\
13	0.143505333653543\\
};
\addlegendentry{Translation Error}

\addplot [color=teal, line width=2.0pt]
  table[row sep=crcr]{%
1	0.0927151528564618\\
2	0.0927151528564618\\
3	0.0252904422252165\\
4	0.07054028487312\\
5	0.0217942009358083\\
6	0.0419271070786404\\
7	0.0222113989291\\
8	0.0657801486652296\\
9	0.082856314959491\\
10	0.0708690878781884\\
11	0.107954651644881\\
12	0.0442865991260302\\
13	0.0368816551243274\\
};
\addlegendentry{Orientation Error}

\end{axis}

%\begin{axis}[%
%width=1.227\figurewidth,
%height=1.227\figureheight,
%at={(-0.16\figurewidth,-0.135\figureheight)},
%scale only axis,
%xmin=0,
%xmax=1,
%ymin=0,
%ymax=1,
%axis line style={draw=none},
%ticks=none,
%axis x line*=bottom,
%axis y line*=left,
%legend style={legend cell align=left, align=left, draw=white!15!black}
%]
%\end{axis}
\end{tikzpicture}%

%% file: Tex/experiment3.tex
\begin{tikzpicture}

\begin{axis}[%
width=0.951\figurewidth,
height=\figureheight,
at={(0\figurewidth,0\figureheight)},
scale only axis,
x dir=reverse,
xmode=log,
xmin=700,
xmax=40000,
xminorticks=true,
xlabel style={font=\color{white!15!black}},
xlabel={Number of Observed Poses},
ymin=0,
ymax=3,
ylabel style={font=\color{white!15!black}},
ylabel={Averaged Errors (mm/degree)},
axis background/.style={fill=white},
title style={font=\bfseries},
title={Density of Poses},
axis x line*=bottom,
axis y line*=left,
xmajorgrids,
xminorgrids,
ymajorgrids,
legend style={at={(0.03,0.97)}, anchor=north west, legend cell align=left, align=left, draw=white!15!black}
]
\addplot [color=purple, line width=2.0pt]
  table[row sep=crcr]{%
40236	0.201521263480447\\
20132	0.145014364980669\\
13440	0.539180800489045\\
10080	0.25946960320589\\
8064	0.134786111180691\\
6720.00000000001	0.214485036453401\\
5768	0.15161161382048\\
5068	0.109504969804356\\
4508.00000000001	0.347193697912948\\
4060.00000000001	0.147628175828914\\
3696	0.252199564103358\\
3388	0.166686804944524\\
3108	0.26732733778103\\
2912	0.377350067954307\\
2716	0.34931352825177\\
2548	0.286784448975441\\
2380	0.432696676966047\\
2268	0.309165694081387\\
2156	0.348357049638215\\
2044	0.363287503453199\\
1932	0.525066187895022\\
1848	0.345496096455859\\
1764	0.454852065298249\\
1708	0.238386550762043\\
1624	0.393504959192128\\
1568	0.455545820058675\\
1512	0.300947959927183\\
1456	0.414207090955274\\
1428	0.212162263650621\\
1372	0.345217778460077\\
1316	0.258892684521809\\
1288	0.303229934517605\\
1260	0.232422308825708\\
1204	0.189992070703501\\
1176	0.21585567080263\\
1148	0.216679522756094\\
1120	0.223094998338781\\
1092	0.372451376509959\\
1064	0.156810113322536\\
1036	0.513152005557648\\
1008	0.532712900306305\\
980.000000000001	0.317098616945843\\
952.000000000001	0.225155811454552\\
952.000000000001	0.279259936920861\\
924	0.221779357609701\\
896.000000000001	0.259138827234578\\
896.000000000001	0.506050706334025\\
868.000000000001	0.653755518840143\\
840.000000000001	0.360452754400807\\
840.000000000001	0.516207663748446\\
812	0.273554486528202\\
812	0.268671249975481\\
784.000000000001	0.238899498461995\\
784.000000000001	0.521711539415138\\
756	0.46115274173517\\
756	0.325606068086383\\
728.000000000001	0.580974525457431\\
728.000000000001	0.320235923681635\\
700	0.297764422181023\\
700	0.249626969549539\\
672.000000000001	0.428791861655134\\
};
\addlegendentry{Translation Error}

\addplot [color=teal, line width=2.0pt]
  table[row sep=crcr]{%
40236	0.0509253485355048\\
20132	0.039950859333775\\
13440	0.0651286045256851\\
10080	0.0650753057297226\\
8064	0.115593387764269\\
6720.00000000001	0.0668003872438776\\
5768	0.114165067443598\\
5068	0.0634403740371905\\
4508.00000000001	0.128440303738394\\
4060.00000000001	0.0626716720357585\\
3696	0.0606402716307297\\
3388	0.0762274501862956\\
3108	0.0523001165850117\\
2912	0.0530477822027651\\
2716	0.0454127813599081\\
2548	0.0697594609557521\\
2380	0.063907318295656\\
2268	0.0764976941381459\\
2156	0.067670603151706\\
2044	0.092178654718631\\
1932	0.107400864203986\\
1848	0.0676237337246546\\
1764	0.0313981833753969\\
1708	0.0410037407513411\\
1624	0.0203458115735708\\
1568	0.0796423973774614\\
1512	0.073149201091776\\
1456	0.102720884545723\\
1428	0.0278711185039038\\
1372	0.100262469993572\\
1316	0.0721805056846021\\
1288	0.038461104342038\\
1260	0.0768197806803252\\
1204	0.0594849803155197\\
1176	0.0628957645482071\\
1148	0.0500631843937791\\
1120	0.069166522295145\\
1092	0.0493027244719482\\
1064	0.0435580376695928\\
1036	0.0740697692014818\\
1008	0.0796265962352098\\
980.000000000001	0.033580821139469\\
952.000000000001	0.0835662769752696\\
952.000000000001	0.0822213340085609\\
924	0.0685124618462538\\
896.000000000001	0.060562993168439\\
896.000000000001	0.0448263189743487\\
868.000000000001	0.0584720356154627\\
840.000000000001	0.0463765974862582\\
840.000000000001	0.065841667340016\\
812	0.0609287107596357\\
812	0.0687211251502449\\
784.000000000001	0.10187647869147\\
784.000000000001	0.039893497860028\\
756	0.0576294040497025\\
756	0.0771332106892748\\
728.000000000001	0.0324804090163759\\
728.000000000001	0.048223602233648\\
700	0.021361966621714\\
700	0.076544504821241\\
700	0.0310845209767088\\
672.000000000001	0.0900820640452222\\
};
\addlegendentry{Orientation Error}

\addplot [color=orange, line width=2.0pt]
  table[row sep=crcr]{%
467.091371682411	0.387813019230246\\
812.848484848485	0.356004057632989\\
1219.27272727273	0.334606091571485\\
1625.69696969697	0.320208617290246\\
2032.12121212121	0.309469078783088\\
2438.54545454546	0.300962170575749\\
3251.39393939394	0.288012331288185\\
4064.24242424243	0.278352630220235\\
5283.51515151516	0.267408675531431\\
6502.78787878789	0.259053312418718\\
8128.48484848485	0.250364873463917\\
10160.6060606061	0.241967837737149\\
13005.5757575758	0.233006060452185\\
16663.3939393939	0.224342720887628\\
21134.0606060606	0.216337410754716\\
26824	0.208594177231915\\
40236	0.196056423716652\\
};
\addlegendentry{Trend Line}

\end{axis}

\begin{axis}[%
width=1.227\figurewidth,
height=1.227\figureheight,
at={(-0.16\figurewidth,-0.135\figureheight)},
scale only axis,
xmin=0,
xmax=1,
ymin=0,
ymax=1,
axis line style={draw=none},
ticks=none,
axis x line*=bottom,
axis y line*=left,
legend style={legend cell align=left, align=left, draw=white!15!black}
]
\end{axis}
\end{tikzpicture}%

%% file: 05-Discussion.tex
\section{DISCUSSION AND CONCLUSION}
\label{sec:conclusion}
In this work, we present a generic and radiation free approach for calibrating a rigidly mounted sensor to the rotation center of a C-arm device based on prior knowledge of the mechanical structure of the device and sensor tracking information using an extended pivot calibration concept. %We call it C-arm pivot calibration problem.  
%In contrast to prior approaches, no radiation is emitted using our approach. 
The proposed radiation-free approach can be applied to any C-arm system configurations as the calibration result defines the relation between the rotational center of the C-arm device and the attached sensor. Given the translation from the rotation center to the X-Ray image plane, the calibration result can be directly used for 2D-2D augmentation.  For CBCT-camera augmentation, if the device has an isocentric design, the rotation center itself is the CBCT reconstruction center; therefore, the calibrated pose from sensor coordinates to C-arm rotation center can also be directly used for AR applications. However, for non-isocentric C-arm devices, the reconstruction center is a virtual scan center instead of the rotation center.  During CBCT sequence acquisition, the position of the gantry is continually adjusted (i.e. the rotation center is moved) such that the center ray subtends the virtual scan center for each scan. In order to use the calibration result for AR applications, the relation between the calibrated rotation center and the virtual scan center has to be recovered.  Note that the relation between the gantry moving trajectory and the virtual center is an essential part of the non-isocentric CBCT volume reconstruction algorithm. Using this relation, we can conduct similar analysis (by adding an offset to the given trajectory) to deduce the relation between the rotation center trajectory and the virtual scan center.  Then, using the calibration result, we can observe rotation center poses during an acquisition and fit the trajectory so as to find its relation to the virtual center.  In a word, the proposed calibration approach is generic and applicable to currently available C-arm devices, given that the X-Ray image or CBCT volume formation is related to the C-arm rotation center.
%the rotation center does not correspond to the center of reconstruction. These systems are characterized by a movement around a virtual scan center instead of the rotation center of the C-arm gantry.  
%In this case, the calibrated center will not remain fixed in the c-circle plane, but will also describe another planar movement. To recover the reconstruction center of the acquired 3D volume, the relationship of the movement in the c-circle plane to the actual rotation center has to be recovered. There are simple movement protocols for which the distance from the source to the virtual scan center remains fixed during CBCT acquisition. In this case, the relationship between the rotation center and the virtual reconstruction center can be recovered with another circle fit in the c-circle plane. However, for more sophisticated movements of robotized non-isocentric C-arm models, the relationship has to be recovered by analyzing the trajectory which needs to be provided by the manufacturer of the system. 
%This observation enables to estimate the offset from calibrated rotation center to virtual reconstruction center for non-isocentric mobile X-Ray imaging systems to close the loop for the application of our method to a generic mobile C-arm. 

%Instead of solving the C-arm pivot calibration problem algebraically by minimizing the energy in Eq.~\ref{eq:pivotenergy} or geometrically by fitting the shape parameterized in Eq.~\ref{eq:etorus}, we decomposed it into several 2D geometry sub-problems and conducted experiments using simulated data.  
Instead of solving the C-arm pivot calibration problem using Eq.~\ref{eq:etorus} or Eq.~\ref{eq:pivotenergy}, we decomposed it into several 2D geometry fitting sub-problems and conducted experiments using simulated data. Our results show 
%that the proposed method is robust to noise, insufficient data and outliers by the application of RANSAC. Furthermore, it can be observed
that the pose observing coordinate system does not influence the calibration result which supports the general applicability of the calibration method. The method is generally robust against noise and outliers in the pose measurements due to the application of RANSAC and is able to give correct results with a minimal set of observed trajectories and is therefore a fast and accurate method. In addition, the final experiment indicates that the algorithm's requirement for the density of observed poses is sufficient to be used with state-of-the-art tracking systems. % In the future, we would like to conduct more throughout comparison between our method and solving directly using Eq.~\ref{eq:pivotenergy} and Eq.~\ref{eq:etorus}.

The proposed concept is not only useful for AR applications such as the overlay of the patient anatomy with the reconstructed 3D volume acquired by CBCT reconstruction for surgical guidance and navigation during orthopedic interventions, but also an important tool to parametrize the C-arm movement space.  Eq.~\ref{eq:etorus} describes the two rotation movements (generalized formula can be derived for the z-axis rotation, base movements, and position adjustment).  It gives a better solution space for applications such as pose initialization for 2D/3D registration of intra-operatively acquired X-Ray images to the pre-operative CT volume including the plan of drill trajectories and cutting planes, or intra-operative panoramic X-Ray image stitching.  For example, Eq.~\ref{eq:etorus} can be used as a constraint when solving for a delta rotational changes. Furthermore, the Torus parametrization could potentially be used as a domain for interpolating X-Ray imaging intrinsic and extrinsic parameters in the whole movement range of the imaging system. 

%Furthermore, we also note that the concept is not limited to C-arm pivot calibration where the pivot locus is a circle.  It works on any known locus.  For instance, instead of performing generic hand-eye calibration on randomly sampled poses, if we are able to control the movement, we should constrain the poses on a known pivot locus and solve for a smaller problem.  Note that the generic hand-eye calibration can be viewed as algebraically solving a pivot calibration on an arbitrary locus.  We believe, by exploiting known geometry, one can solve an easier and more well-defined problem, which should yield for better calibration results.

% In future work, we will evaluate this novel calibration method with different tracking systems, as well as camera types and show the robustness and applicability of our method to C-arm models with isocentric and non-isocentric design. We believe that the parametrization of the C-arm movement as a Torus shape will enable new concepts for parameter mapping to C-arm movements. Possible concepts include the refinement of tracking results by mapping the observed poses back to the Torus surface, utilizing the calibrated shape as domain for a full mapping of X-Ray imaging intrinsic and extrinsic camera parameters by inter- and extrapolating between calibrated poses. Furthermore, we want to use the Torus shape as a domain for sonification parameter mapping for C-arm movement which could facilitate intuitive re-positioning of X-Ray views during orthopedic procedures.

In future work, we will further evaluate this novel method with different real-world tracking systems and sensors, and show its robustness and applicability to different C-arm models including both isocentric and non-isocentric design. We furthermore remark that the proposed concept is not limited to C-arm pivot calibration where the pivot locus is a circle.  By exploiting a known geometry for pivoting, the complexity of the problem can be reduced which leads to a more robust solution and better results.% This parameterization can potentially be used for C-arm tracking refinement by mapping the observed poses back to the parameterized surface, and for utilizing the calibrated Torus as a domain for interpolating X-Ray imaging intrinsic and extrinsic parameters in the whole movement range of the imaging system. 

% \keymessage{Example of application also includes replacing hand-eye calibration by defining known geometry at a joint, then our method can be applied to calibrate the translation to the joint. <-- this can be an insight of our geometry-based formulation.}